\documentclass[conference,a4paper]{IEEEtran}
\IEEEoverridecommandlockouts
\usepackage{cite}
\usepackage{amsmath,amssymb,amsfonts}
\usepackage{algorithmic}
\usepackage{colortbl}
\usepackage{graphicx}
\usepackage{textcomp}
\usepackage{xcolor}
\usepackage{xspace}
\usepackage{mathtools}
\usepackage{multirow}
\usepackage{url}
\usepackage{comment}

\usepackage{subcaption}
\usepackage[acronyms,nonumberlist,nopostdot,nomain,nogroupskip]{glossaries}
\usepackage{booktabs}
\usepackage{hyperref}
\newacronym{6g}{6G}{sixth generation}
\newacronym{3gpp}{3GPP}{3rd Generation Partnership Project}
\newacronym{adc}{ADC}{Analog to Digital Converter}
\newacronym{dac}{DAC}{Digital to Analog Converter}
\newacronym{5g}{5G}{5th generation}
\newacronym{aimd}{AIMD}{Additive Increase Multiplicative Decrease}
\newacronym{am}{AM}{Acknowledged Mode}
\newacronym{amc}{AMC}{Adaptive Modulation and Coding}
\newacronym{aoa}{AoA}{Angle of Arrival}
\newacronym{aod}{AoD}{Angle of Departure}
\newacronym{aqm}{AQM}{Active Queue Management}
\newacronym{awgn}{AGWN}{Additive White Gaussian Noise}
\newacronym{balia}{BALIA}{Balanced Link Adaptation}
\newacronym{bdp}{BDP}{Bandwidth-Delay Product}
\newacronym{bf}{BF}{Beamforming}
\newacronym{fpga}{FPGA}{field-programmable gate array}
\newacronym{cc}{CC}{Congestion Control}
\newacronym{cdf}{CDF}{Cumulative Distribution Function}
\newacronym{cn}{CN}{Core Network}
\newacronym{cm}{CM}{confusion matrix}
\newacronym[plural=\gls{cnn}s,firstplural=convolutional neural networks (CNNs)]{cnn}{CNN}{convolutional neural network}
\newacronym{cqi}{CQI}{Channel Quality Information}
\newacronym{cp}{CP}{Control Plane}
\newacronym{csirs}{CSI-RS}{Channel State Information - Reference Signal}
\newacronym{dc}{DC}{Dual Connectivity}
\newacronym{dce}{DCE}{Direct Code Execution}
\newacronym{dci}{DCI}{Downlink Control Information}
\newacronym{dmr}{DMR}{Deadline Miss Ratio}
\newacronym{dmrs}{DMRS}{DeModulation Reference Signal}
\newacronym{e2e}{E2E}{End-to-End}
\newacronym{ecn}{ECN}{Explicit Congestion Notification}
\newacronym{ebs}{EBS}{exhaustive beam sweep}
\newacronym{edf}{EDF}{Earliest Deadline First}
\newacronym{enb}{eNB}{evolved Node Base}
\newacronym{epc}{EPC}{Evolved Packet Core}
\newacronym{es}{ES}{Edge Server}
\newacronym{fdma}{FDMA}{Frequency Division Multiple Access}
\newacronym{fdd}{FDD}{Frequency Division Duplexing}
\newacronym[firstplural=Radio Access Technologies (RATs)]{rat}{RAT}{Radio Access Technology}
\newacronym{fs}{FS}{Fast Switching}
\newacronym{txer}{TX}{transmitter}
\newacronym{rxer}{RX}{receiver}
\newacronym{bt}{BT}{beam tracking}
\newacronym{ftp}{FTP}{File Transfer Protocol}
\newacronym{gnb}{gNB}{Next Generation Node Base}
\newacronym{bs}{BS}{Base Station}
\newacronym{harq}{HARQ}{Hybrid Automatic Repeat reQuest}
\newacronym{hetnet}{HetNet}{Heterogeneous Network}
\newacronym{hh}{HH}{Hard Handover}
\newacronym{hol}{HOL}{Head-of-Line}
\newacronym{ia}{IA}{initial access}
\newacronym{imt}{IMT}{International Mobile Telecommunication}
\newacronym{iot}{IoT}{Internet of Things}
\newacronym{los}{LOS}{Line-of-Sight}
\newacronym{lte}{LTE}{Long Term Evolution}
\newacronym{m2m}{M2M}{Machine to Machine}
\newacronym{ml}{ML}{machine learning}
\newacronym{dl}{DL}{deep learning}
\newacronym{mac}{MAC}{Medium Access Control}
\newacronym{mc}{MC}{Multi-Connectivity}
\newacronym{mcs}{MCS}{Modulation and Coding Scheme}
\newacronym{mec}{MEC}{Mobile Edge Cloud}
\newacronym{mi}{MI}{Mutual Information}
\newacronym{mimo}{MIMO}{Multiple Input, Multiple Output}
\newacronym{mmwave}{mmWave}{millimeter wave}
\newacronym{mmWave}{mmWave}{Millimeter wave}
\newacronym{mptcp}{MPTCP}{Multipath TCP}
\newacronym{mr}{MR}{Maximum Rate}
\newacronym{mss}{MSS}{Maximum Segment Size}
\newacronym{mtd}{MTD}{Machine-Type Device}
\newacronym{mtu}{MTU}{Maximum Transmission Unit}
\newacronym{nfv}{NFV}{Network Function Virtualization}
\newacronym{nlos}{NLOS}{Non-Line-of-Sight}
\newacronym{nr}{NR}{New Radio}
\newacronym{ofdm}{OFDM}{Orthogonal Frequency Division Multiplexing}
\newacronym{pdcch}{PDCCH}{Physical Downlink Control Channel}
\newacronym{pdcp}{PDCP}{Packet Data Convergence Protocol}
\newacronym{pdsch}{PDSCH}{Physical Downlink Shared Channel}
\newacronym{pdu}{PDU}{Packet Data Unit}
\newacronym{pf}{PF}{Proportional Fair}
\newacronym{pgw}{PGW}{Packet Gateway}
\newacronym{phy}{PHY}{Physical}
\newacronym{pbch}{PBCH}{Physical Broadcast Channel}
\newacronym[plural=\gls{mme}s,firstplural=Mobility Management Entities (MMEs)]{mme}{MME}{Mobility Management Entity}
\newacronym{prb}{PRB}{Physical Resource Block}
\newacronym{pss}{PSS}{Primary Synchronization Signal}
\newacronym{pucch}{PUCCH}{Physical Uplink Control Channel}
\newacronym{pusch}{PUSCH}{Physical Uplink Shared Channel}
\newacronym{rach}{RACH}{Random Access Channel}
\newacronym{ran}{RAN}{Radio Access Network}
\newacronym{red}{RED}{Random Early Detection}
\newacronym{rf}{RF}{Radio Frequency}
\newacronym{rlc}{RLC}{Radio Link Control}
\newacronym{rlf}{RLF}{Radio Link Failure}
\newacronym{rrc}{RRC}{Radio Resource Control}
\newacronym{rrm}{RRM}{Radio Resource Management}
\newacronym{rr}{RR}{Round Robin}
\newacronym{rs}{RS}{Remote Server}
\newacronym{rsrp}{RSRP}{Reference Signal Received Power}
\newacronym{rss}{RSS}{Received Signal Strength}
\newacronym{rtt}{RTT}{Round Trip Time}
\newacronym{rw}{RW}{Receive Window}
\newacronym{rx}{RX}{Receiver}
\newacronym{sa}{SA}{standalone}
\newacronym{sack}{SACK}{Selective Acknowledgment}
\newacronym{sap}{SAP}{Service Access Point}
\newacronym{ap}{AP}{Access Point}
\newacronym{sch}{SCH}{Secondary Cell Handover}
\newacronym{scoot}{SCOOT}{Split Cycle Offset Optimization Technique}
\newacronym{sdma}{SDMA}{Spatial Division Multiple Access}
\newacronym{sinr}{SINR}{Signal to Interference plus Noise Ratio}
\newacronym{sm}{SM}{Saturation Mode}
\newacronym{snr}{SNR}{Signal-to-Noise-Ratio}
\newacronym{son}{SON}{Self-Organizing Network}
\newacronym{ss}{SS}{Synchronization Signal}
\newacronym{ssbs}{SSBs}{synchronization signal blocks}
\newacronym{ssb}{SSB}{synchronization signal block}
\newacronym{srs}{SRS}{Sounding Reference Signal}
\newacronym{sss}{SSS}{Secondary Synchronization Signal}
\newacronym{tb}{TB}{Transport Block}
\newacronym{tcp}{TCP}{Transmission Control Protocol}
\newacronym{tdd}{TDD}{Time Division Duplexing}
\newacronym{tdma}{TDMA}{Time Division Multiple Access}
\newacronym{tfl}{TfL}{Transport for London}
\newacronym{tm}{TM}{Transparent Mode}
\newacronym{trp}{TRP}{Transmitter Receiver Pair}
\newacronym{tti}{TTI}{Transmission Time Interval}
\newacronym{ttt}{TTT}{Time-to-Trigger}
\newacronym{tx}{TX}{Transmitter}
\newacronym{ue}{UE}{User Equipment}
\newacronym{ul}{UL}{Uplink}
\newacronym{uml}{UML}{Unified Modeling Language}
\newacronym{um}{UM}{Unacknowledged Mode}
\newacronym{utc}{UTC}{Urban Traffic Control}
\newacronym{vm}{VM}{Virtual Machine}
\newacronym{rsrq}{RSRQ}{Reference Signal Received Quality}
\newacronym{rssi}{RSSI}{Received Signal Strength Indicator}
\newacronym{crs}{CRS}{Cell Reference Signal}
\newacronym{nsa}{NSA}{Non Stand Alone}
\newacronym{mrdc}{MR-DC}{Multi \gls{rat} \gls{dc}}
\newacronym{endc}{EN-DC}{E-UTRAN-\gls{nr} \gls{dc}}
\newacronym{5gc}{5GC}{5G Core}
\newacronym{si}{SI}{Study Item}
\newacronym{iab}{IAB}{Integrated Access and Backhaul}
\newacronym{wf}{WF}{Wired-first}
\newacronym{hqf}{HQF}{Highest-quality-first}
\newacronym{pa}{PA}{Position-aware}
\newacronym{mlr}{MLR}{Maximum-local-rate}
\newacronym{wbf}{WBF}{Wired Bias Function}
\newacronym{mib}{MIB}{Master Information Block}
\newacronym{sib}{SIB}{Secondary Information Block}
\newacronym{kpi}{KPI}{Key Performance Indicator}
\newacronym{ppp}{PPP}{Poisson Point Process}
\newacronym{gtp}{GTP}{GPRS Tunneling Protocol}
\newacronym{amf}{AMF}{Access and Mobility Management Function}
\newacronym{dash}{DASH}{Dynamic Adaptive Streaming over HTTP}
\newacronym{http}{HTTP}{HyperText Transfer Protocol}
\newacronym{qos}{QoS}{Quality of Service}
\newacronym{udp}{UDP}{User Datagram Protocol}
\newacronym{cu}{CU}{Central Unit}
\newacronym{du}{DU}{Distributed Unit}
\newacronym{mt}{MT}{Mobile Termination}
\newacronym{sdap}{SDAP}{Service Data Adaptation Protocol}
\newacronym{tdm}{TDM}{Time Division Multiplexing}
\newacronym{fdm}{FDM}{Frequency Division Multiplexing}
\newacronym{sdm}{SDM}{Space Division Multiplexing}
\newacronym{dag}{DAG}{Directed Acyclic Graph}
\newacronym{st}{ST}{Spanning Tree}
\newacronym{ummimo}{UM-MIMO}{Ultra-massive Multiple Input, Multiple Output}
\newacronym{uavs}{UAVs}{Unmanned Aerial Vehicles}
\newacronym{wlan}{WLAN}{Wireless LAN}
\newacronym{rlnc}{RLNC}{Random Linear Network Coding}
\newacronym{drx}{DRX}{Discontinuous Reception}
\newacronym{cpu}{CPU}{Central Processing Unit}
\newacronym{txb}{TXB}{transmitter's beam}
\newacronym{rxb}{RXB}{receiver's beam}
\newacronym{sifs}{SIFS}{Short Interframe Space}
\newacronym{difs}{DIFS}{DCF Interframe Space}

\newacronym{rfid}{RFID}{Radio Frequency Identification}
\newacronym{rfp}{RFP}{radio fingerprinting}
\newacronym{sdr}{SDR}{software-defined radio}
\newacronym{dnn}{DNN}{deep neural network}
\def\highlight{\bf \cellcolor{lightgray}}
\newcommand{\FRAMEWORK}{\texttt{BottleFit}\xspace}

\def\BibTeX{{\rm B\kern-.05em{\sc i\kern-.025em b}\kern-.08em
    T\kern-.1667em\lower.7ex\hbox{E}\kern-.125emX}}
\begin{document}

\title{\FRAMEWORK: Learning Compressed Representations in Deep Neural Networks for Effective and Efficient Split Computing
\thanks{This work was supported by the Intel Corporation and the NSF grant MLWiNS-2003237, CNS-2134973 and CNS-2120447.}
}

\author{\large Yoshitomo Matsubara$^\dagger$, Davide Callegaro$^\dagger$, Sameer Singh$^\dagger$, Marco Levorato$^\dagger$, and Francesco Restuccia$^*$\\
\normalsize $^\dagger$Department of Computer Science, University of California, Irvine, United States\\
  $^*$Institute for the Wireless Internet of Things, Northeastern University, United States\\
\normalsize E-mails: \{yoshitom, dcallega, sameer, levorato\}@uci.edu, frestuc@northeastern.edu}




\maketitle

\begin{abstract}
Although mission-critical applications require the use of \glspl{dnn}, their continuous execution at mobile devices results in a significant increase in energy consumption. While edge offloading can decrease energy consumption, erratic patterns in channel quality, network and edge server load can lead to severe disruption of the system's key operations.
An alternative approach, called \textit{split computing}, generates compressed representations within the model (called ``bottlenecks''), to reduce bandwidth usage and energy consumption. Prior work has proposed approaches that introduce additional layers,  to the detriment of energy consumption and latency. For this reason, we propose a new framework called \textit{\FRAMEWORK}, which, in addition to targeted DNN architecture modifications, includes  a novel training strategy to achieve high accuracy even with strong compression rates. We apply \FRAMEWORK on cutting-edge \gls{dnn} models in image classification, and show that \FRAMEWORK achieves 77.1\% data compression with up to 0.6\% accuracy loss on ImageNet dataset, while state of the art such as SPINN loses up to 6\% in accuracy. We experimentally measure the power consumption and latency of an image classification application running on an NVIDIA Jetson Nano board (GPU-based) and a Raspberry PI board (GPU-less). We show that \FRAMEWORK decreases power consumption and latency respectively by up to 49\% and 89\% with respect to (w.r.t.) local computing and by 37\% and 55\% w.r.t. edge offloading. We also compare \FRAMEWORK with state-of-the-art autoencoders-based approaches, and show that (i) \FRAMEWORK reduces power consumption and execution time respectively by up to 54\% and 44\% on the Jetson and 40\% and 62\% on Raspberry PI; (ii) the size of the head model executed on the mobile device is 83 times smaller. We publish the code repository for reproducibility of the results in this study.
\end{abstract}

\glsresetall
\IEEEpeerreviewmaketitle

\section{Introduction}
\label{sec:intro}

Emerging mobile applications, such as real-time navigation and obstacle avoidance on drones~\cite{kouris2018learning,kukkala2018advanced,smolyanskiy2017toward}, increasingly require the execution of complex inference tasks. The key challenge in supporting these applications is that state-of-the-art \gls{dnn} models are characterized by computational requirements that go far beyond what the vast majority of mobile devices can offer today.
Recent trends approach this problem by (\emph{i}) reducing model complexity using knowledge distillation (KD)~\cite{hinton2014distilling} and model pruning/quantization~\cite{jacob2018quantization}, and (\emph{ii}) designing lightweight models such as MobileNet~\cite{sandler2018mobilenetv2} and MnasNet~\cite{tan2019mnasnet}, at the expense of significant accuracy loss, \emph{e.g.}, up to 6.4\% compared to the large \gls{cnn} model ResNet-152~\cite{he2016deep}.

An alternative approach is edge computing~\cite{zhou2018robust,chen2019deep,xie2019source}, where the execution is completely offloaded to edge servers. While in some scenarios high-throughput links are possible -- \emph{e.g.}, a 5G link in Line of Sight (LoS) -- the channel quality may suddenly fluctuate, thus impairing performance. For instance, protocol interactions triggered by mobility and impaired propagation have been proven to induce erratic capacity patterns even in high-capacity 5G links~\cite{zhang2019will}
. Moreover, most  \gls{iot} devices do not support high data rates, and instead rely on lower-power technologies such as WiFi and LoRa~\cite{samie2016iot}, the latter having maximum data rate of 37.5 kbps due to duty cycle and other limitations \cite{adelantado2017understanding}.

\begin{figure}[t]
    \centering
    \includegraphics[width=0.95\linewidth]{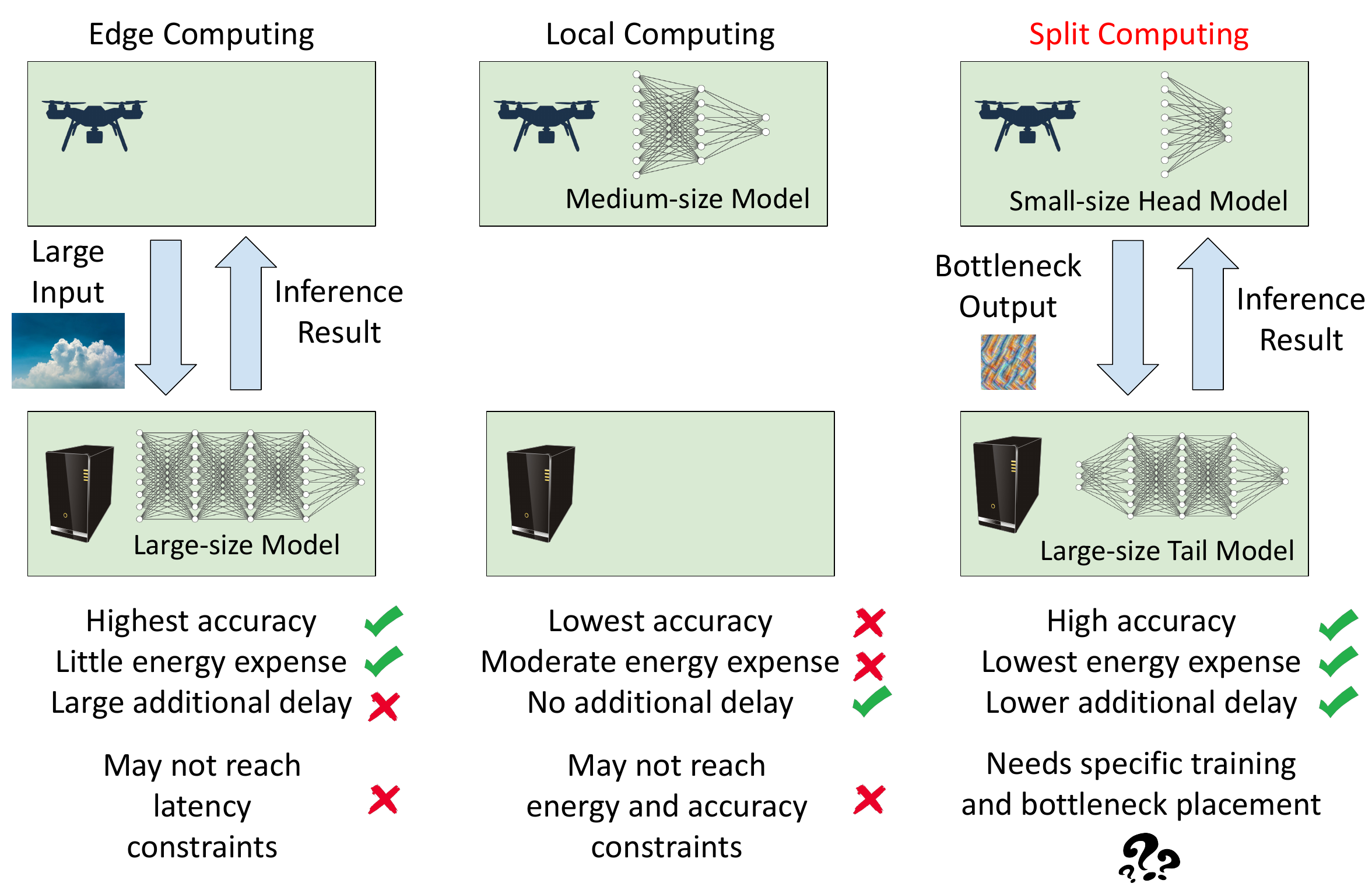}
    \vspace{-0.5em}
    \caption{Pros and cons of edge, local and split computing.}
    \label{fig:intro}
\end{figure}

The challenging edge-device dilemma can be tackled with \emph{split computing}~\cite{kang2017neurosurgeon}, which is depicted in Fig.~\ref{fig:intro}. The idea is to split a larger \gls{dnn} into head and tail models executed at the mobile device and edge server, respectively. In the most advanced splitting strategies, the original model is modified by introducing a ``bottleneck'' layer~\cite{eshratifar2019bottlenet,matsubara2019distilled,hu2020fast,shao2020bottlenet++,jankowski2020joint,matsubara2020head,matsubara2020neural}. The core intuition behind this strategy is to let the part of the model before the bottleneck learn a \textit{compressed} feature representation. This way, the output tensor of the head model -- which is designed to be smaller than the input size -- is transmitted over the channel to the edge server instead of the input data. The compressed representation is then used by the tail network to produce the final prediction output (\emph{e.g.}, classification), which is sent back to the mobile device.
Notice that different from federated learning~\cite{konevcny2016federated}, training is performed offline (\emph{e.g.}, at edge/cloud servers), while split computing occurs at runtime.
\smallskip

\textbf{Challenges.}~The success of split computing critically hinges on its capability to substantially decrease energy and latency with respect to traditional computational paradigms. \textit{In other words, the key issue is to design bottleneck architectures that can achieve the best trade-off between head model compression and tail model accuracy.} This problem is extremely challenging, since the introduction of a bottleneck with high compression rate usually comes to the cost of severe detriment in accuracy. The key innovation, therefore, would be to design custom-tailored training strategies to maintain high accuracy despite the bottleneck compression. 
 
We summarize below the core contributions of this paper.

$\bullet$ We present \FRAMEWORK, a novel framework for split computing. The key innovation of \FRAMEWORK is a novel multi-stage training strategy to achieve high accuracy even with strong compression rates. In short, in the first stage we pretrain \emph{encoder} and \emph{decoder} structures built around the bottleneck to mimic the intermediate representations in the original pretrained model (without bottlenecks), while freezing parameters in the subsequent layers. Then, we perform a refinement stage where we adapt the compressed representation learnt in the first stage to a target task; 

$\bullet$ We apply \FRAMEWORK on cutting-edge \glspl{cnn} such as DenseNet-169, DenseNet-201 and ResNet-152 on the ImageNet dataset, and compare the accuracy obtained by \FRAMEWORK with state-of-the-art local computing~\cite{sandler2018mobilenetv2} and split computing approaches~\cite{eshratifar2019bottlenet,matsubara2019distilled,hu2020fast,matsubara2020head,matsubara2020split,shao2020bottlenet++}.  Our training campaign concludes that \FRAMEWORK achieves up to 77.1\% data compression (with respect to JPEG) with only up to 0.6\% loss in accuracy, while existing mobile and split computing approaches incur considerable accuracy loss of up to 6\% and 3.6\%, respectively. For the sake of completeness, we compare \FRAMEWORK with an autoencoder-based approach, which loses up to 16\% in accuracy with respect to \FRAMEWORK. 

$\bullet$  We experimentally measure the power consumption and latency of an image classification application. We consider two configurations (i) a Jetson Nano board (GPU-based) communicating with the edge server (a high performance Laptop) using WiFi and Long-Term Evolution, and (ii) a Raspberry PI board (GPU-less) communicating using LoRa.  We show that \FRAMEWORK decreases power consumption and latency respectively by up to 49\% and 89\% with respect to (w.r.t.) local computing and by 37\% and 55\% w.r.t. edge offloading. Notably, our head model is 83 times smaller compared to state-of-the-art autoencoder-based approaches. 

$\bullet$ Finally, we release the code repository of \FRAMEWORK and trained model weights produced in this study to ensure reproducibility.\footnote{\url{https://github.com/yoshitomo-matsubara/bottlefit-split_computing}}

\section{Related Work}
\label{sec:related_work}

Many different aspects of mobile edge computing have been extensively investigated~\cite{zhou2019predictive,li2020mvstylizer}.
The key assumption of edge-based computing is that wireless links are stable and high capacity, which is seldom the case in many practical scenarios. For this reason, split computing was introduced to divide the computation between the mobile device and the edge~\cite{kang2017neurosurgeon}.
Early contributions, such as~\cite{kang2017neurosurgeon}, split the neural model while leaving unaltered its architecture. This approach has been shown to grant limited improvement over traditional local or edge computing in most application scenarios. Building on this framework, other contributions introduce compression at the splitting point~\cite{choi2018deep} and consider autoencoders and generative models to provide computing options alternative to local and edge computing~\cite{eshratifar2019jointdnn}. SPINN~\cite{laskaridis2020spinn} improves the performance Neurosurgeon by introducing ``early exits'' in existing models, but also incurs a significant deradation of the original accuracy in complex tasks.

In this paper, we focus on bottleneck-injected models to realize effective and efficient split computing. Most of the existing  studies train the altered models from scratch~\cite{eshratifar2019bottlenet,hu2020fast,shao2020bottlenet++}. Others reuse pretrained parameters in available architectures for the tail model, while re-designing and re-training the head portion to introduce a bottleneck~\cite{matsubara2019distilled,matsubara2020neural,matsubara2020head}. These latter contributions introduce the notion of Head Network Distillation (HND) and Generalized HND (GHND), that use knowledge distillation in the training process.
None of these studies provide a comprehensive discussion on training methodologies, and the bottleneck-injected models are either not assessed~\cite{emmons2019cracking} or assessed in relatively simple classification tasks~\cite{eshratifar2019bottlenet,matsubara2019distilled,hu2020fast,jankowski2020joint,shao2020bottlenet++} such as miniImageNet, Caltech 101, CIFAR-10 and -100 datasets. For example, CIFAR datasets present an input RGB image of $32 \times 32$ pixels, that is, transferring the image would require a very short time and the compression granted by split computing strategies is likely unnecessary. Furthermore, many modern applications require higher resolution images and more complex classes. In this paper, we use the large-scale ImageNet dataset~\cite{russakovsky2015imagenet} for image classification,\footnote{\url{https://image-net.org/}} where the most common input tensor size for CNN models is $3 \times 224 \times 224$, which gives us a strong motivation to compress tensors for split computing.

Some recent architectures make use of autoencoders at the splitting point~\cite{yao2020deep}. We thoroughly compare our approach, where we directly modify the layers of the model to embed a bottlenck, with those frameworks in Section~\ref{subsec:ae_injection} and show that \FRAMEWORK greatly improves accuracy. We note that the introduction of autoencoders in the models increases the overall computational complexity. For instance, compared to DeepCOD~\cite{yao2020deep} our head model is 83 times smaller in the worst case. This leads to a reduction in energy consumption at the mobile device of up to 54\% (on a Jetson Nano) and 44\% (Raspberry Pi 4), as well as lower encoding time -- up to 86\% less on the Jetson Nano and 82\% on the Raspberry Pi 4 -- and overall execution time.

\newcommand{\norm}[1]{\left\lVert #1 \right\rVert}

\section{\FRAMEWORK: Proposed Framework}
\label{sec:framework}

We first introduce the system model in Section~\ref{subsec:sys_mod}, describe the bottleneck design in Section~\ref{subsec:bottleneck}, then we present our new training strategy in Section~\ref{subsec:training}, and finally illustrate the performance tradeoff guiding our system design in Section~\ref{subsec:placement}.

\subsection{System Model and Preliminaries}
\label{subsec:sys_mod}

\begin{figure}[t]
    \centering
    \includegraphics[width=\linewidth]{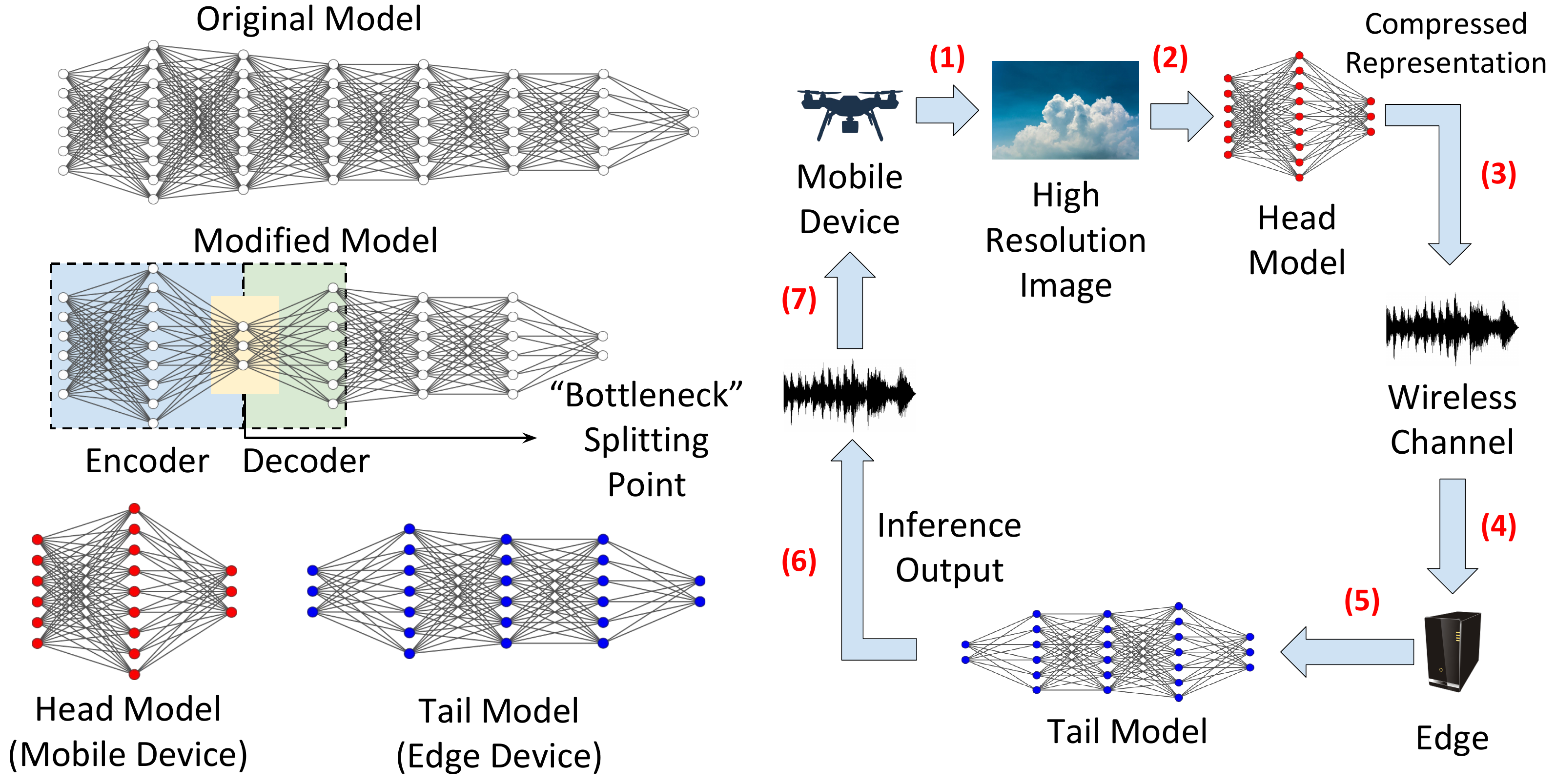}
    \vspace{-1.5em}
    \caption{(left) Split computing: the original model is redesigned with a ``bottleneck'' and then split into head and tail sections. (right) Example of split computing system. Note that the training process is not split but done offline.}
    \label{fig:overview}
\end{figure}

The core idea behind \emph{split computing} is to divide a \gls{dnn} into \textit{head} and \textit{tail} models, which are executed on the mobile device and the edge server at runtime, respectively.
The left side of Fig.~\ref{fig:overview} shows an example of split computing with bottlenecks, while the right side shows a concrete example of a split computing system. When split computing is used in our setting, the mobile device captures high-resolution images (step 1), which are then fed to the head model, ultimately tasked to produce a compressed representation by its output tensor (step 2).
The result is transmitted over the wireless channel (step 3) and received by the edge server (step 4).
The compressed representation, then, is fed to the tail model to produce the inference output such as predicted class label (step 5), which is sent over the wireless channel (step 6) and received by the mobile device (step 7). 

In the following, the notation $\mathcal{R}$ indicates the set of real numbers. Without loss of generality, we consider a deep neural network (DNN) model defined as a mapping $f(\mathbf{x}_i; \mathbf{\theta}) : \mathcal{R}^{i} \rightarrow \mathcal{R}^{o}$ of an input representation $\mathbf{x}_i \in \mathcal{R}^{i}$ to an output representation $\mathbf{x}_o \in \mathcal{R}^{o}$.
By defining as $L$ the number of layers, the DNN mapping is computed through $L$ subsequent transformations given a model input $\mathbf{x}$:
\begin{equation}
    \mathbf{o}_{j} = \left\{
    \begin{array}{ll}
        \mathbf{x}  & j = 0 \\
        f_j(\mathbf{o}_{j-1}, \mathbf{\theta}_j) \hspace{1cm} & 1 \le j \le L  
    \end{array}
    \right.
    \label{eq:dnn_layers}
\end{equation}
\noindent where $\mathbf{o}_j$ and $\mathbf{\theta} = \{\theta_1, \ldots, \theta_L\}$  are the output of the $j$-th layer and the set of parameters of the DNN. Note that $f_j$ can be a low-level layer (\emph{e.g.}, convolution, pooling and fully-connected layers), but also a high-level layer consisting of multiple low-level layers such as residual block or ``shortcut'' in ResNet models~\cite{he2016deep}, and dense block in DenseNet models~\cite{huang2017densely}.

\subsection{Bottleneck Design}
\label{subsec:bottleneck}
To build and define the bottlenecks, we introduce \emph{encoder} and \emph{decoder} structures within an original pretrained model. Specifically, we replace the first $l_\text{ed}$ layers in the original model with an encoder and decoder. The former structure is positioned from the first layer to the bottleneck, and plays a role of ``compressor'', generating a compact tensor from the input sample. The latter is composed of the layers as part of the tail model that decompress the encoded object, \emph{i.e.}, the bottleneck's output to recreate the output of an intermediate layer. \textit{We point out that while traditional autoencoders compress and reconstruct an input, we modify the layers to operate in an encoder-decoder fashion, which (i) maps the model input to an intermediate output and (ii) is trained to execute a given downstream task without excessive loss in accuracy. We show in Section \ref{subsec:ae_injection} that autoencoders lose about 16\% in accuracy with respect to our approach.} 

Clearly, the design of the encoder/decoder (\emph{e.g.}, position in the model, dimension of the bottleneck) influences the tradeoff between computing load at the mobile device, overall complexity and compression gain, which ultimately determines key performance metrics such as energy consumption, delay and accuracy.
 Thus, when introducing bottlenecks, we need to carefully (i) design the encoder and decoder; (ii) choose the bottleneck placement in the head model; (iii) preserve the accuracy of the unaltered original model as much as possible. The following sections address all these aspects to build the \FRAMEWORK framework.

\begin{figure}[h]
    \centering
    \includegraphics[width=0.99\linewidth]{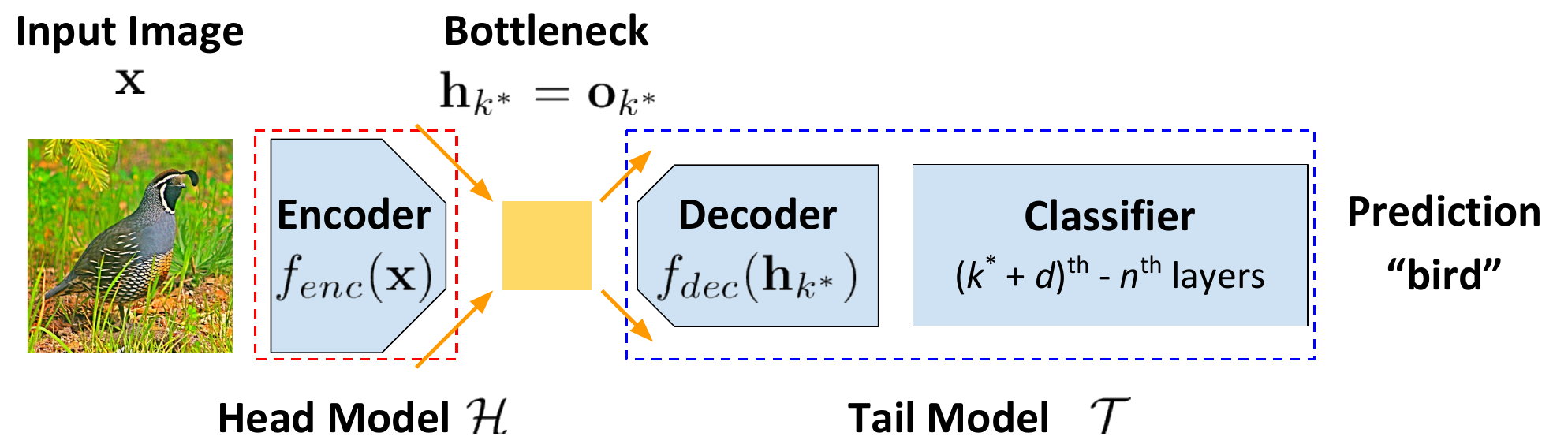}
    \vspace{-1em}
    \caption{Model components: encoder, decoder and classifier. Note that the last component, classifier, consists of the last $(n - (k^* + d))$ layers in the original pretrained model.}
    \label{fig:components}
\end{figure}

\textbf{Bottleneck modeling:}~Given an original pretrained model consisting of $N$ layers, we design and introduce bottlenecks to the model, and retrain the bottleneck-injected model to preserve accuracy as much as possible.
A bottleneck-injected model is composed of $n<N$ layers and takes as input a tensor whose shape is identical to that for the original model.
As illustrated in Fig.~\ref{fig:components}, the head model $\mathcal{H}$ and tail model $\mathcal{T}$ are built by splitting the model at the bottleneck layer $\mathcal{B}$ (=$f_{k^*}$).
The head model $\mathcal{H}$ overlaps with the encoder $f_\text{enc}$, composed of $k^*$ layers \emph{i.e.},
\begin{eqnarray}
    \mathcal{H} = f_\text{enc} = \left\{
    \begin{array}{ll}
        \mathbf{h}_{0} = \mathbf{o}_{0} = x  \\
        \mathbf{h}_{j} = f_j(\mathbf{o}_{j-1}, \theta_j)  & 1 \le j \le k^* - 1, \\
        \mathbf{h}_{k^*}  = \mathcal{B}(\mathbf{o}_{k^* - 1}, \theta_{\mathcal{B}})&  
    \end{array}
    \right.
    \label{eq:head_model}
\end{eqnarray}
\noindent where $\theta_{\mathcal{B}}$ is the set of parameters of the bottleneck layer $\mathcal{B}$, and $\mathbf{h}_{k^*}$ indicates the bottleneck representation to be transferred from the mobile device to the edge server in inference session.

The bottleneck representation $\mathbf{h}_{k^*}$ is then fed to the tail model $\mathcal{T}$, which consists of the decoder $f_\text{dec}$ ($d$ layers where $l_\text{ed} = k^* + d$) and the remaining ($n - l_\text{ed}$) layers in the original model.
\begin{equation}
    \mathcal{T} = \left\{
    \begin{array}{ll}
        f_\text{dec} = \left\{
        \begin{array}{ll}
            \mathbf{t}_{0} = \mathbf{h}_{k^*}  &  \\
            \mathbf{t}_{j - k^* + 1} = f_j(\mathbf{o}_{j-1}, \theta_j)& k^* < j \le l_\text{ed}
        \end{array}
        \right.\\
        \mathbf{t}_{j - k^* + 1} = f_j(\mathbf{o}_{j-1}, \theta_j) \hspace{5em} l_\text{ed} < j \le n 
    \end{array}
    \right.
    \label{eq:tail_model}
\end{equation}

\textit{We remark that different from traditional autoencoders, we do not create additional layers for the bottleneck, thus the overall complexity of a bottleneck-injected model will not increase from that of the original pretrained model.} Instead, to introduce bottlenecks we modify the head portion of the original pretrained models that are often overparameterized.\smallskip

\textbf{Encoder-Decoder Implementation:} To obtain the bottleneck, we then define a new sequence of encoder and decoder layers.
Specifically, given a sequence of the first layers in the original pretrained model, we design encoder-decoder architectures using the following steps:

\noindent 1) Decompose high-level layers (\emph{e.g.}, dense blocks in DenseNet \cite{huang2017densely} and residual blocks in ResNet \cite{he2016deep}) in the sequence into low-level layers such as convolution, batch normalization and ReLU layers;

\noindent 2) Prune a subset of the layers and define a new sequence by the $l_\text{ed}$ remaining layers. If necessary, a set of new layers can be added: deconvolution layers for upsampling after the bottleneck point and/or pooling layers for better convergence;
    
    \noindent 3) Determine the location of the bottleneck in the new sequence ($k^*$-th layer in the $l_\text{ed}$ layers);
    
    \noindent 4) Adjust layer-specific hyperparameters of layers such as number of output channels, kernel size and kernel-stride to have the sequence's output shape match that expected by the remaining layers in the original pretrained model.

We use convolution layers to create the bottleneck at the $k^*$-th layer ($1 \le k^* \le l_\text{ed}$) since convolution layers allow us to control their output tensor shape with respect to channel, height and width.
Then, we choose 2 consecutive convolution layers in the new sequence to build bottlenecks $\mathbf{o}_\mathcal{B} = \mathbf{o}_{k^*}$ at the $k^*$-th layer defined in the previous section, and the following layers gradually decode the compressed representation. To achieve further compression, we also reduce patch size (height and width) of the bottleneck representation. We use a slightly larger kernel-stride size in the early convolution layer(s) to reduce all the output channels, width and height of the output tensor, and introduce a deconvolution layer after the bottleneck layer so that the decoded representation can match the tensor shape expected by the following layers.

\subsection{Multi-Stage Training Strategy}
\label{subsec:training}

Our core intuition to preserve accuracy is to maximize the performance of the \emph{encoding} and \emph{decoding} capabilities of the layers neighboring the splitting point to produce an output preserving the overall task performance. Specifically: (\emph{i}) the training strategy should be sophisticated enough to train low-complexity encoders $f_\text{enc}$ and (\emph{ii}) the following modules including the decoder should adapt the compressed representations to the downstream tasks.

\begin{figure}[t]
    \centering
    \includegraphics[width=0.9\linewidth]{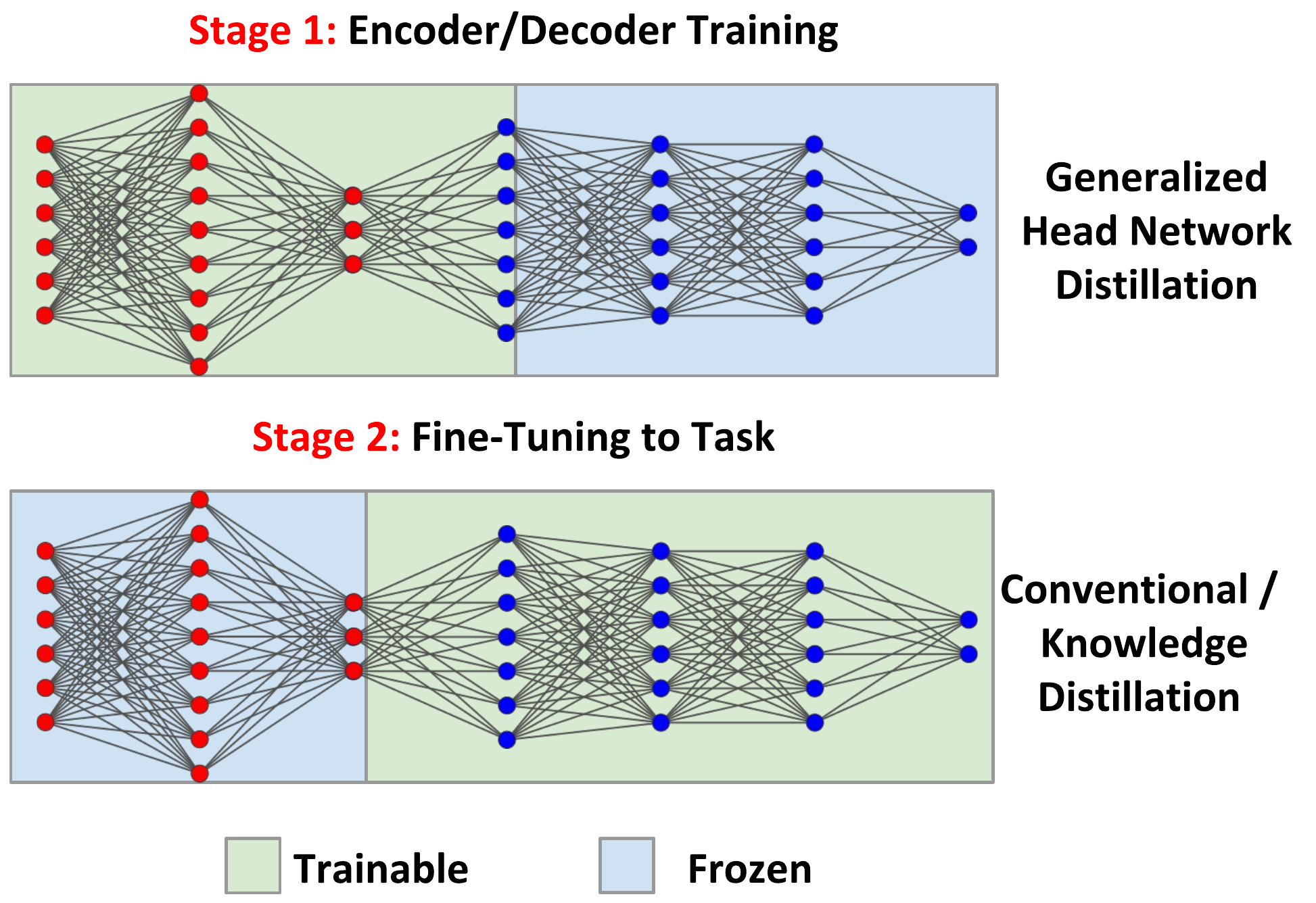}
    \caption{Proposed multi-stage training method. It is optional to freeze the encoder's parameters in the stage 2.}

    \label{fig:training}
\end{figure}

Figure~\ref{fig:training} illustrates our proposed multi-stage training method for bottleneck-injected models.
We focus on the training of the compressed bottleneck representations, and adapt the learnt representations to the target task, which in this study is image classification on the ImageNet dataset.
In the following, we will refer to the original and modified models as \textit{teacher} and \textit{student} models, respectively.\smallskip

{\bf Stage 1 -- Pretraining Encoder-Decoder:} This stage focuses on training both the encoder $f_\text{enc}$ and decoder $f_\text{dec}$ in the modified model to reconstruct the representations of the corresponding layer in the original model. At this stage, we use a loss function -- Generalized Head Network Distillation (GHND) -- which considers the output of multiple layers in the model:
\begin{equation}
    \mathcal{L}_\text{Pre}(x) = \lambda_\text{ed} || t_\text{ed}(x) - f_\text{dec}(f_\text{enc}(x)) ||_{2}^{2} + \sum_{j \in \mathcal{J}} \lambda_{j} || t_{j}(x) - s_{j}(x) ||_{2}^{2},
    \label{eq:ghnd_loss}
\end{equation}
\noindent
where $t_\text{ed}$ denotes a function consisting of teacher's layers to be replaced with the encoder $f_\text{enc}$ and decoder $f_\text{dec}$ in the student model. For example, the function $t_\text{ed}$ for DenseNet-169, DenseNet-201 and ResNet-152 consist of layers until (including) the 2nd transition layer in DenseNet-169 and -201~\cite{huang2017densely}, and the 2nd block in ResNet-152~\cite{he2016deep}, respectively. ${j}$ is a loss index for a pair of layers that are components of classifier in teacher and student models (\emph{i.e.}, these layers are frozen and in neither encoder nor decoder), and $t_{j}$ and $s_{j}$ denote teacher and student model functions of input data $x$ for the loss index $j$ (\emph{i.e}, intermediate outputs of layers in teacher and student models), respectively.\footnote{For simplicity, we define $s_{j}$ as a nested function using the first $K_j$ layers in student model \emph{i.e.}, $s_{j}(x) = f_{K_j}(f_{K_{j}-1}(...(f_{1}(x))))$, and the same applies to teacher model.}
$\lambda_{*}$ is a balancing factor and set to 1 in this study. Importantly, we consider the outputs of frozen layers, in addition to those from the trainable layers during the head network distillation process.




{\bf Stage 2 -- Adapting Bottleneck to the Target Task:} In this stage, we fine-tune the remaining components of the model to suppress reconstruction errors produced by the encoder-decoder pair while optionally freezing the parameters of the encoder as illustrated in Fig.~\ref{fig:training}. The classifier learns representations for the downstream task adopting the bottleneck representations learnt to reconstruct the output of the teacher's head model.


We test different approaches to fine-tune the models with encoder-decoder we pretrained in the 1st stage.
One could simply use a conventional fine-tuning (FT) method that minimizes a task-specific loss $\mathcal{L}_\text{task}$, \emph{e.g.}, a standard cross entropy loss $\mathcal{L}_\text{CE}$ defined in Eq.~\ref{eq:ce_loss} for classification
\begin{equation}
    \mathcal{L}_\text{CE}(x, y) = -\log \left( \frac{\exp \left( s(x, y) \right)}{\sum_{j \in \mathcal{C}} \exp\left( s(x, j) \right)} \right),
    \label{eq:ce_loss}
\end{equation}
\noindent where $x$ and $y$ indicate an input image and the associated class label respectively. $\mathcal{C}$ is a set of class labels ($|\mathcal{C}| = 1,000$ for the ImageNet dataset), and $s(x, j)$ is a model's predicted value for a class $j$ given an input image $x$. We can use knowledge distillation (KD) instead of the conventional fine-tuning (FT) approach, then minimize the following loss:
\begin{equation}
    \mathcal{L}_\text{KD}(x, y) = \alpha \mathcal{L}_\text{task}(x, y) + (1 - \alpha) \tau^2  KL \left( q(x), p(x) \right),
    \label{eq:kd_loss}
\end{equation}
\noindent
where $\alpha$ is a balancing factor (hyperparameter) between \emph{hard target} (left term) and \emph{soft target} (right term) losses.
$\mathcal{L}_\text{task}$ is a task-specific loss function, and it is a cross entropy loss in image classification tasks \emph{i.e.}, $\mathcal{L}_\text{task} = \mathcal{L}_\text{CE}$.
$KL$ is the Kullback-Leibler divergence function, where $p(x)$ and $q(x)$ are probability distributions of teacher (\emph{soft-target}) and student models for an input $x$, that is, $p(x) = [p_{1}(x), \cdots, p_{|\mathcal{C}|}(x)]$ and $q(x) = [q_{1}(x), \cdots, q_{|C|}(x)]$:
\begin{equation}
    p_{c}(x) = \frac{\exp \left( \frac{t(x, c)}{\tau} \right)}{\sum_{j \in \mathcal{C}} \exp\left( \frac{t(x, j)}{\tau} \right)}, ~~q_{c}(x) = \frac{\exp \left( \frac{s(x, c)}{\tau} \right)}{\sum_{j \in \mathcal{C}} \exp\left( \frac{s(x, j)}{\tau} \right)},
    \label{eq:pq}
\end{equation}
\noindent
where $t(x, j)$ indicates a teacher model's predicted value for a class $j$ given a resized input image $x$, and $\tau$ is a ``temperature'' (hyperparameter) for knowledge distillation that is normally set to 1. We set $\alpha$ to 0.5 as in ~\cite{hinton2014distilling}.

\subsection{Latency-Power-Accuracy Trade-Off}
\label{subsec:placement}
We break down the end-to-end delay $D$ as the sum of the following random variables: (i) the head model execution time $D_{\mathcal{H}}(k^*, f^*)$, (ii) communication time $D_{\mathcal{N}}(k^*, f^*)$, and (iii) tail model execution time $D_{\mathcal{T}}(k^*, f^*)$.
The power consumption at the mobile device is modeled as the sum of the head model execution $P_{\mathcal{H}}(k^*, f^*)$ and communication $P_{\mathcal{N}}(k^*, f^*)$ components.
The distributions of these random variables are dependent on the particular mobile device and wireless technology, as well as the model design, and will be computed experimentally.
For simplicity, in the following we will drop the $k^*$ and $f^*$ notations. The end-to-end delay then is $D_\text{e2e} = D_{\mathcal{H}} + D_{\mathcal{N}} + D_{\mathcal{T}}$, while the power consumption at the mobile device is  $P_\text{md} = P_{\mathcal{H}} + P_{\mathcal{N}}$. Intuitively, the design of the model also influences the attainable accuracy.
The tradeoff between these measures guide our design and can be used to find the best configuration given system-level parameters.


\begin{table*}[t]
\caption{Validation accuracy [\%]$^{\dagger}$ of models with bottlenecks trained on ImageNet dataset by \underline{baseline methods}.}
\vspace{-1.5em}
\begin{center}
    \bgroup
    \setlength{\tabcolsep}{0.55em}
    \def\arraystretch{1.1}
    \small
    \begin{tabular}{l|rrr|rrr}
    \toprule
    \multicolumn{1}{c|}{Bottleneck} & \multicolumn{3}{c|}{12 output channels} & \multicolumn{3}{c}{3 output channels}\\ \midrule
	\multicolumn{1}{c|}{Base Model} & \multicolumn{1}{c|}{DenseNet-169} & \multicolumn{1}{c|}{DenseNet-201} & \multicolumn{1}{c|}{ResNet-152} & \multicolumn{1}{c|}{DenseNet-169} & \multicolumn{1}{c|}{DenseNet-201} & \multicolumn{1}{c}{ResNet-152} \\ \midrule
	\multicolumn{1}{c|}{Conventional~\cite{eshratifar2019bottlenet,hu2020fast,shao2020bottlenet++}} & 66.90 ~(-8.700) & 68.92 ~(-7.970) & 72.02 ~(-11.04) & 60.69 ~(-14.91) & 62.85 ~(-14.04) & 66.86 ~(-11.45) \\
    \multicolumn{1}{c|}{KD~\cite{hinton2014distilling}} & 69.37 ~(-6.230) & 70.89 ~(-6.000) & 74.06 ~(-4.250) & \highlight 62.66 ~(-12.94) & \highlight 64.09 ~(-12.80) & \highlight 67.61 ~(-10.71) \\
	\multicolumn{1}{c|}{HND~\cite{matsubara2020head}} & \highlight 72.03 ~(-3.570) & \highlight 73.62 ~(-3.270) & \highlight 75.13 ~(-3.180) & 55.18 ~(-20.48) & 56.57 ~(-20.32) & 53.40 ~(-24.91) \\ \bottomrule
	\end{tabular}
	\egroup
\end{center}
\footnotesize \raggedright $\dagger$ ImageNet (ILSVRC 2012) test dataset is not publicly available. $\ddagger$ Numbers in brackets indicate difference from the original models.
\label{table:baseline_results}
\end{table*}

\section{\FRAMEWORK Accuracy Evaluation}
\label{sec:model_assess}

In this section, we describe baseline methods and compare their performance with \FRAMEWORK.
Specifically, we will consider  ImageNet (ILSVRC 2012)~\cite{russakovsky2015imagenet}, a popular large-scale benchmark dataset for image classification.
For ResNet-152, we design two splitting points, called {\bf Splitting Point 1} (SP1) and {\bf Splitting Point 2} (SP2).

\vspace{0.5mm}
\noindent
\textbf{Splitting Point 1 (SP1):}
We design an early bottleneck by reducing the output channels of the 2nd convolution layer.
This way, we obtain a bottleneck representation whose patch size is $29 \times 29$ for each of the output channels defined in the 2nd convolution layer \emph{e.g.}, $12 \times 29 \times 29$ if the 2nd convolution layer has 12 output channels.
Besides the tensor shape, the actual size of the bottleneck representation to be transferred will be explained in Section~\ref{subsec:eval_classification}. 

\vspace{0.5mm}
\noindent
\textbf{Splitting Point 2 (SP2):}
SP2 is injected in the third convolution layer. The encoder-decoder architecture we embed (\emph{e.g.}, in ResNet-152) is designed to mimic the output representation shaped $256 \times 28 \times 28$. Using a convolution layer, we can reduce the number of channels at bottleneck point from 256 channels to 12 or fewer channels. If we reduce the width and height ($28 \times 28$ in this case) at the bottleneck point (\emph{e.g.}, $14 \times 14$), however, we need then to upsample the compressed tensor with respect to height and width ($14 \times 14$ to $28 \times 28$) by a deconvolution layer so that the decoder output matches the input tensor shape expected by the 2nd block.

We note that later placements in the model would lead to an excessive computing load assigned to the mobile device, which would in turn would result in an increased overall execution time sufficiently large to offset a larger compression gain.

\subsection{Baseline Evaluation}
\label{subsec:baseline_methods}
First, we describe the training strategies used in previous papers that we consider as baseline training methods.

\vspace{0.5mm}
\noindent
\textbf{Conventional Training}
In~\cite{eshratifar2019bottlenet,hu2020fast,shao2020bottlenet++}, the training strategy applied to bottleneck-injected models is the same conventional strategy used for the training of CNN models.
Specifically, training uses cross-entropy (CE) loss defined in Eq. (\ref{eq:ce_loss}) as guiding metric with learning rate adjustment.

\vspace{0.5mm}
\noindent
\textbf{Knowledge Distillation:}
We also consider knowledge distillation (KD)~\cite{hinton2014distilling} as a baseline approach used in the previous study~\cite{matsubara2020head}.
In KD, the whole model is treated as a ``student model'' and trained with the original pretrained model (teacher) by minimizing a linear combination of soft-labeled and hard-labeled losses as shown in Eq. (\ref{eq:kd_loss}).

\vspace{0.5mm}
\noindent
\textbf{Head Network Distillation:}
In (HND) ~\cite{matsubara2019distilled,matsubara2020head}, the the head bottleneck-injected models (student) are trained with the output of the original head model (teacher).
This approach does not need human-annotated data (\emph{e.g.}, class label) for training encoder and decoder as the training target is the output from the teacher head model. 


Table~\ref{table:baseline_results} shows the baseline performance on ImageNet dataset~\cite{russakovsky2015imagenet} by the above three training methods. We empirically find that such bottleneck-injected models trained by the manners used in their studies suffer from degraded accuracy in a complex task, and the accuracy loss is more significant when introducing very small bottlenecks.
Furthermore, we confirm that models trained by KD consistently outperform those trained using the conventional training method. Notably, another consistent trend is that models with bigger bottlenecks trained by HND outperform those trained by the other methods, but KD performs best for those with 4 times smaller bottlenecks. 
We can see that the performance difference between the conventional and KD methods is small compared to that between the HND and KD (or conventional) methods.
Interestingly, the performance gap between HND and KD for models with smaller bottlenecks (3 output channels) is, instead, quite perceivable. This implies that compressing bottleneck representations makes it difficult for the models to train with HND, which performs best for those with bigger bottlenecks (12 output channels).

\subsection{\FRAMEWORK Evaluation}
\label{subsec:eval_classification}

We train exactly the same models on ImageNet dataset used in Section~\ref{subsec:baseline_methods}, reusing the same training configurations for a fair comparison. Specifically, the number of training epochs is set to 20, with the first and last 10 epochs are dedicated to the 1st and 2nd stages with Adam and SGD optimizers, respectively.
The initial learning rates are set to 0.001 in both the optimizers, and decayed by 0.1 after the first 5 epochs in each stage.\smallskip

\noindent
{\bf SP1:} We first focus on SP1 and examine the models with the smallest bottleneck (3 output channels) as these models have more room to emphasize the accuracy improvements over the best baseline method for the models (See Table~\ref{table:baseline_results}).
We pretrain encoder and decoder at the 1st stage of training, and use both the conventional and KD methods for the 2nd stage with/without freezing encoder pretrained in the 1st stage, thus there are four configurations of the proposed approach.
As shown in Table~\ref{table:proposed_results_3ch}, all these configurations significantly outperform the best baseline method for models with 3 output channels for bottlenecks, with an accuracy improvement of 4.1 - 5.7 points. Interestingly, the results with the four different configurations are comparable whereas we confirm gaps between the conventional and KD methods in the baseline evaluation (Table~\ref{table:baseline_results}). Given that the proposed multi-stage training methods result in comparable accuracy, we apply the first configuration, Pretraining$\rightarrow$FT (FE) to the models with bigger bottlenecks (6, 9 and 12 output channels) as it requires the least training cost among the four configurations.

\begin{table}[t]
\caption{Validation accuracy [\%]$^{\dagger}$ of models with bottlenecks \underline{(3 output channels)} trained by \underline{our proposed methods}.}
\vspace{-1em}
\begin{center}
    \footnotesize
    \begin{tabular}{l|rrr} \toprule
	\multicolumn{1}{c|}{Base Model} & \multicolumn{1}{c|}{DenseNet-169} & \multicolumn{1}{c|}{DenseNet-201} & \multicolumn{1}{c}{ResNet-152} \\ \midrule
	\multicolumn{1}{c|}{Best Baseline (KD)} & 62.66 & 64.09 & 67.61 \\ \midrule
	\multicolumn{1}{c|}{Pretraining$\rightarrow$FT (FE)} & 68.41 & 69.45 & 71.66 \\
    \multicolumn{1}{c|}{Pretraining$\rightarrow$KD (FE)} & 68.10 & \highlight 69.61 & 71.62 \\
	\multicolumn{1}{c|}{Pretraining$\rightarrow$FT} & \highlight 68.43 & 69.41 & 71.50 \\
	\multicolumn{1}{c|}{Pretraining$\rightarrow$KD} & 68.23 & 69.54 & \highlight 71.73 \\ \bottomrule
	\end{tabular}
\end{center}
\footnotesize \raggedright FE: Frozen encoder pretrained in the 1st stage
\label{table:proposed_results_3ch}
\end{table}

\begin{figure*}[t]
    \begin{center}
        \begin{subfigure}[t]{0.325\linewidth}
            \centering
            \includegraphics[width=1.0\linewidth]{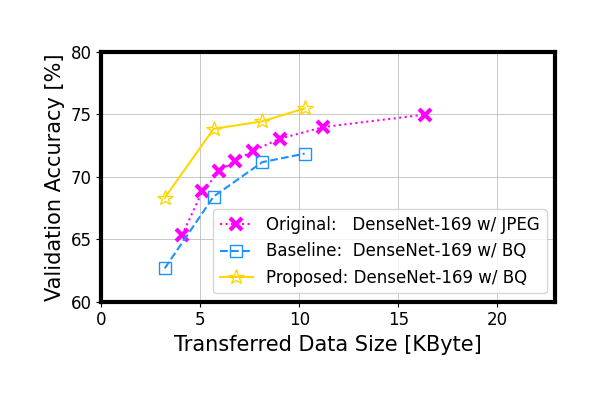}
            \vspace{-3em}
            \caption{DenseNet-169\label{fig:tradeoff-densenet169}}
        \end{subfigure}
        \begin{subfigure}[t]{0.325\linewidth}
            \centering
            \includegraphics[width=1.0\linewidth]{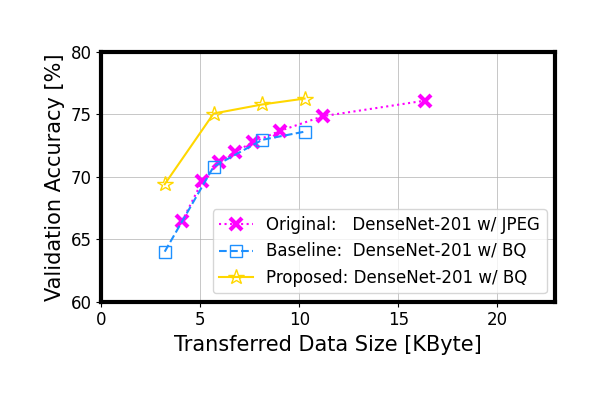}
            \vspace{-3em}
            \caption{DenseNet-201\label{fig:tradeoff-densenet201}}
        \end{subfigure}
        \begin{subfigure}[t]{0.325\linewidth}
            \centering
            \includegraphics[width=1.0\linewidth]{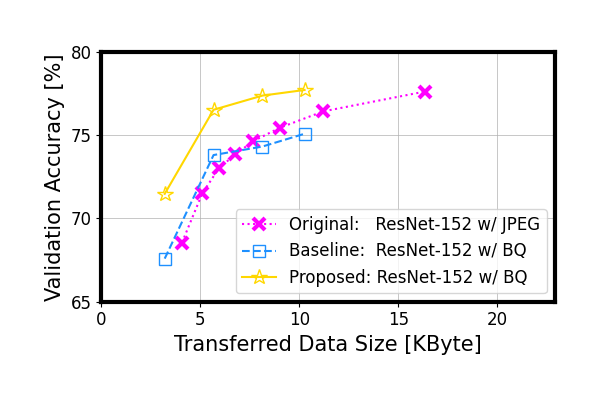}
            \vspace{-3em}
            \caption{ResNet-152\label{fig:tradeoff-resnet152}}
        \end{subfigure}
    \end{center}
    \centering
    \vspace{-1em}
    \caption{Transferred data size vs. accuracy for three different base models. \FRAMEWORK lifts up the baseline performance of bottleneck-injected models and improves over JPEG compression.\vspace{-2em}}
    \label{fig:data_vs_accuracy}
\end{figure*}

\begin{figure}[t]
    \centering
    \includegraphics[width=0.98\linewidth]{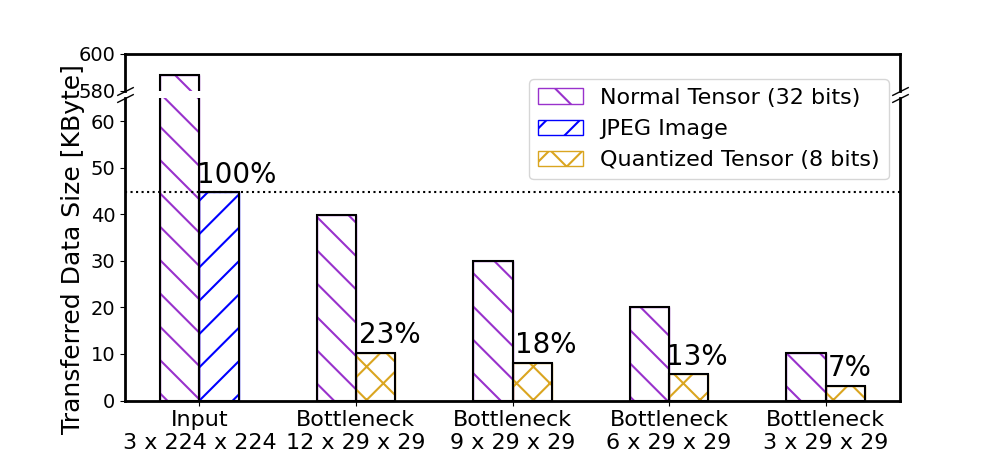}
    \caption{Tensor size and corresponding transferred data size for our DenseNet-169, DenseNet-201, and ResNet-152.}
    \label{fig:tensor_data_size}
\end{figure}

Figure~\ref{fig:data_vs_accuracy} illustrates the trade-off between transferred data size and accuracy. To further compress the size of data to be transferred, we apply a post-training bottleneck quantization (BQ) technique to the output at the bottleneck point.
Specifically, the quantization technique proposed in~\cite{jacob2018quantization} is applied to represent the bottleneck tensor (32-bit floating point) by 8-bit integer tensor and one 32-bit value to dequantize at edge side.
Interestingly, as shown in Fig.~\ref{fig:data_vs_accuracy}, BQ technique does not degrade the accuracy of the model while reducing the transferred data size by approximately 75\%.
In Fig.~\ref{fig:data_vs_accuracy}, we show the best accuracy of each model in Table~\ref{table:baseline_results} as a baseline.
As for the curve with JPEG compression, the transferred data size is average JPEG file size in the ImageNet dataset, resized and center-cropped to $224 \times 224$ pixels for DenseNet-169, DenseNet-201 and ResNet-152.
The data size used to compute the total delay for bottleneck-injected models corresponds to the output from the bottleneck (splitting point).
\textit{For all the considered models, the approach we propose achieves up to 5.6\% improvement in accuracy for each of the models with larger bottlenecks, while the trained models achieve the accuracy of the original pretrained models while reducing 93.3\% of tensor elements to be transmitted.
Furthermore, this improvement of the trade-off has the bottleneck-injected models significantly outperform the original pretrained models with JPEG compression.}
Figure~\ref{fig:tensor_data_size} reports the actual data size of model input and bottleneck output used in the inference time evaluation.
In addition to about 75\% data size reduction of tensor data by bottleneck quantization (32-bit to 8-bit), the quantized bottleneck saves up to 93\% compared to the size of JPEG-compressed model input.



\vspace{1mm}
\noindent
{\bf SP2:} We now discuss the trade-off between bottleneck size and accuracy in SP2, where the bottleneck reduces by up 98\% the amount of transferred data with respect to the input JPEG image.
Figure~\ref{fig:ae_and_bottleneck_place} illustrates a trend of trade-off between bottleneck size and accuracy for the models with bottlenecks introduced to their later layers (SP2).
As expected, while later bottlenecks impose additional computing load, the accuracy vs compression trend improves, that is, the accuracy degradation is smaller given a compression rate, specifically when further compression (leftward in the figure) is required.

\subsection{Comparison with Autoencoders}
\label{subsec:ae_injection}

For a comparison purpose, we consider an autoencoder (AE) consisting of $L_\text{AE}$ layers (= $L_\text{AE}^\text{Enc}$ + $L_\text{AE}^\text{Dec}$ layers for its encoder and decoder) introduced within a pretrained model composed by $L_\text{M}$ layers at the $k_\text{AE}$-th layer ($1 < k_\text{AE} < L_\text{M}$). As discussed in Section~\ref{sec:related_work}, this approach has been widely used in the literature, but has the drawback of increasing models' complexity both of the head and tail. By directly modifying the layers of the models, \FRAMEWORK creates encoder/decoder structures transforming the output of an intermediate layer into that of a later one through a bottleneck, thus embedding compression in the computing process.

\begin{figure}[t]
    \centering
    \includegraphics[width=0.98\linewidth]{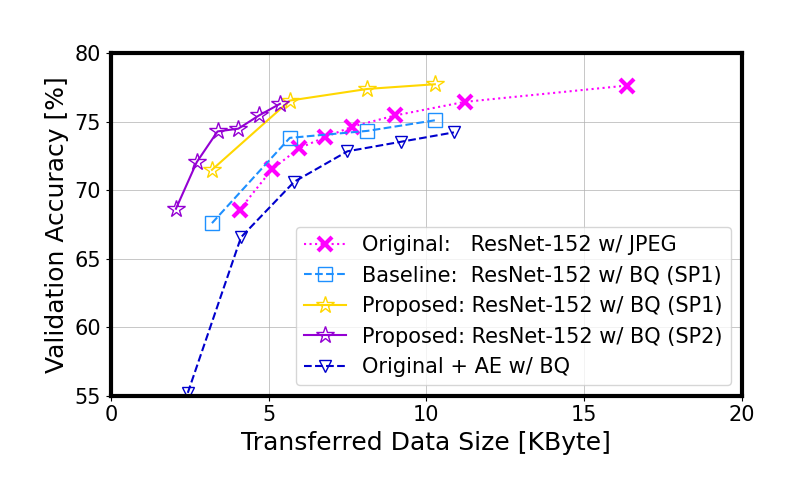}
    \vspace{-1.5em}
    \caption{Comparison of \FRAMEWORK with (i) our ResNet-152 with bottleneck introduced to its later layer (SP2), and (ii) original ResNet-152 with autoencoder to its 2nd block.}
    \label{fig:ae_and_bottleneck_place}
\end{figure}

Then, the first $k_\text{AE}$ and $L_\text{AE}^\text{Enc}$ layers in the original model and the encoder of the AE respectively are executed by the mobile device, and the remaining layers ($L_\text{AE}^\text{Dec}$ and ($L_\text{M} - k_\text{AE}$) layers) are executed to complete inference.
As the additional layers increase computing load, we design lightweight AE using 4 convolution layers (encoder) and 4 deconvolution layers (decoder).
We also use batch normalization and ReLU layers between convolution (and deconvolution) layers for better convergence in training.
We train AEs on ImageNet for 20 epochs with the following hyperparameters: batch size of 32, initial learning rate of 0.001 decayed by a factor of 10 every 5 epochs.
The parameters of AE are learned with Adam optimizer~\cite{kingma2015adam} by minimizing a reconstruct loss (sum of squared loss).
Due to the limited space, we put our focus on ResNet-152 in this experiment, and inject the AE between the 2nd and 3rd residual blocks. Figure~\ref{fig:ae_and_bottleneck_place} shows the resulting trend compared to the baseline and proposed methods.
Interestingly, AEs do not reach even the best baseline methods, while they increase computing load compared to simple and bottleneck-injected model splitting.



\section{Experimental Evaluation}
\label{sec:experimental_results}

In this section, we extensively evaluate \FRAMEWORK through a real-world experimental testbed.
We first describe our experimental setup in Section \ref{sec:exp_setup}, then present the delay and power consumption measurements obtained from Jetson Nano in Section \ref{sec:res_jetson}, and finally the delay and power consumption obtained from Raspberry Pi 4 in Section \ref{sec:res_raspberry}.

\begin{figure}[h]
    \centering
    \includegraphics[width=0.825\linewidth]{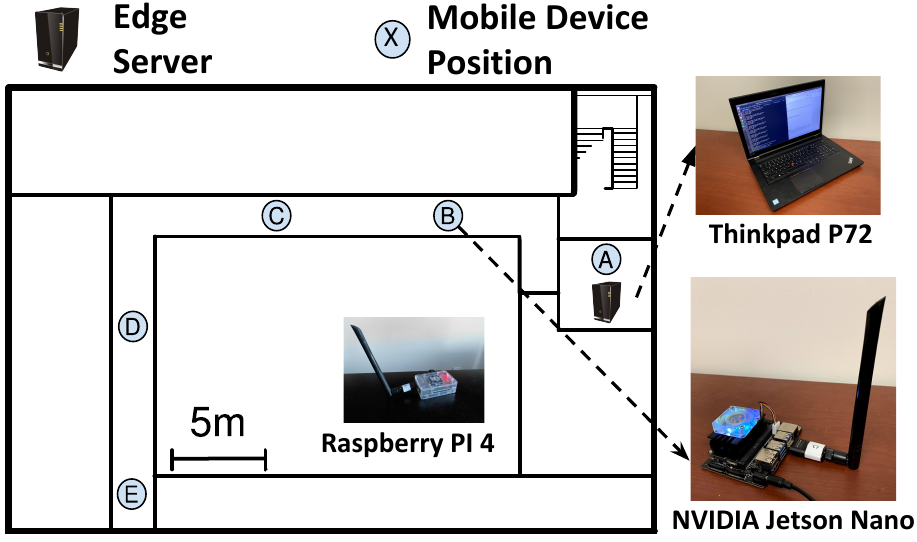}
    \vspace{-0.5em}
    \caption{Experimental setup.\vspace{-1em}}
    \label{fig:floorplan}
\end{figure}

\subsection{Experimental Setup}\label{sec:exp_setup}

The experiments were performed indoor in the Donald Bren Hall at University of California, Irvine as shown in Fig.~\ref{fig:floorplan}. We evaluate \FRAMEWORK on several configurations of embedded platforms and communication technologies. As mobile device, we use either an NVIDIA Jetson Nano, quad-core ARM $1.9$GHz CPU and mounting a $128$-core GPU operating at $0.95$GHz and 4GB of 64-bit LPDDR4 memory or a Raspberry Pi 4, mounting a quad-core ARM $1.5$GHz CPU and $2$GB of LPDDR4 memory. As edge server, we use a ThinkPad P72 with hexa-core CPU operating up to $4.3$GHz, 32GB of memory and NVIDIA GPU Quadro P600 that has $384$ cores operating at $1.45$GHz as edge server.

In terms of communication technologies we use:

\vspace{1.5mm}
\noindent
{\bf Wi-Fi:} We use the laptop's Intel Wireless-AC 9560 as hotspot to which the Jetson Nano connects through a Realtek WiFi dongle supporting IEEE 802.11n. The figure shows the location of the edge server, which acts as WiFi hot-spot, to which Jetson Nano board connects as client. We implemented an application generating $300$ images in total, at a rate of $5$ images/sec. We placed the mobile device in different positions in the building (Fig.~\ref{fig:floorplan}) to evaluate the impact of link quality on performance. For each of these positions we measure the average quality link  -- as provided by the \texttt{ss} tool in Linux. In the following, we will show the average values as well as 90\% confidence intervals.

\vspace{1.5mm}
\noindent
{\bf LoRa:} the Long Range (LoRa) technology is a widely adopted technology used in Internet of Things (IoT) applications. We extract achievable transmission rates from the networking dataset available in \cite{ferran2017lora}, and then, the communication time based on the data size. The total latency is then computed as the sum of the experimental execution time of the models' sections and the communication time.

\vspace{1.5mm}
\noindent
{\bf Long-Term Evolution:} We use the traces reported in \cite{raca2018beyond} to compute the transmission rate when Long-Term Evolution (LTE) is used by the distributed system. The traces have been collected while driving in densely serviced areas.

\subsection{Latency and Power -- Jetson Nano/Wi-Fi and LTE}\label{sec:res_jetson}

\begin{figure}[t!]
    \centering
    \includegraphics[width=0.95\linewidth]{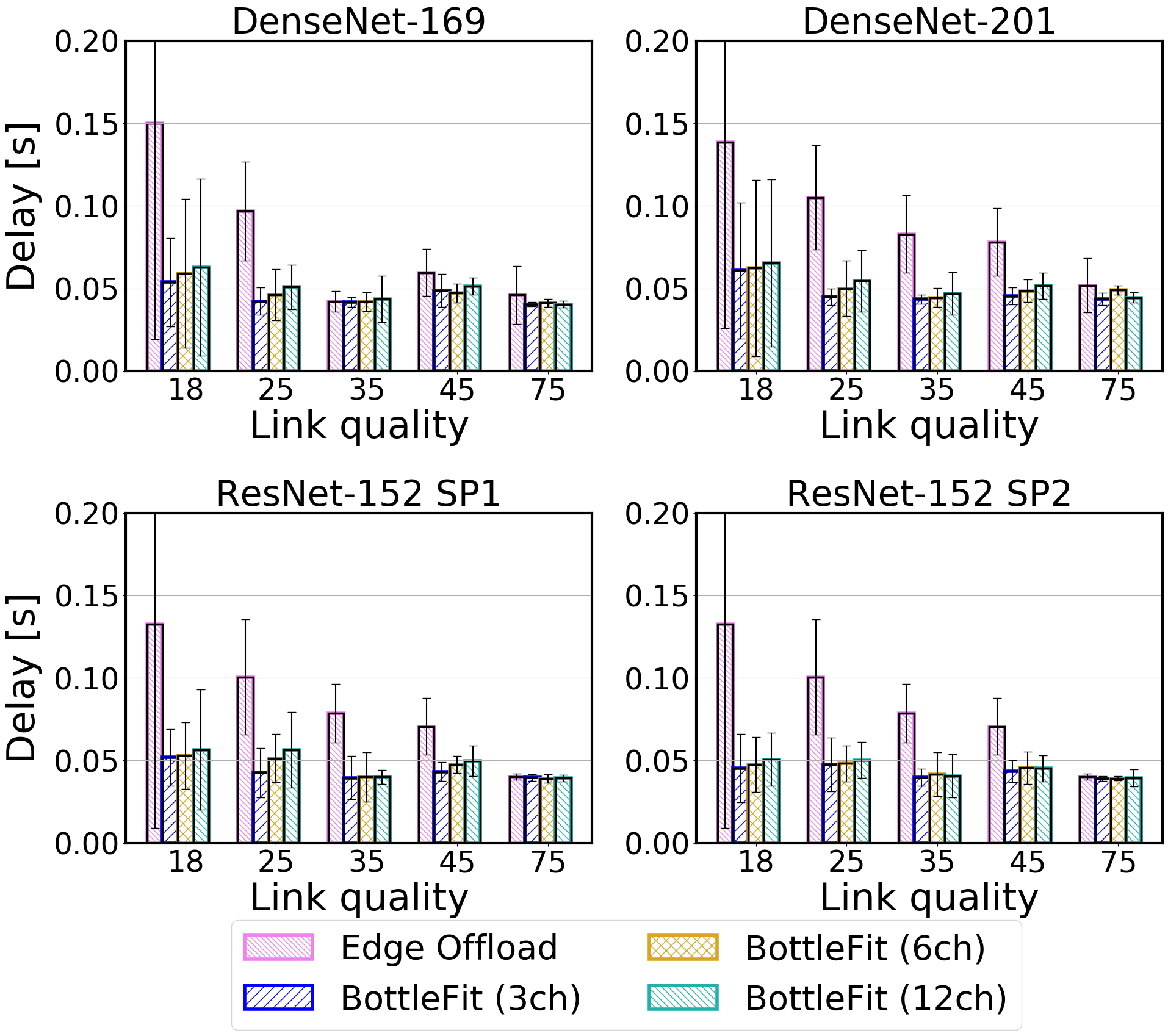}
    \vspace{-0.5em}
    \caption{Delay of the considered models run on \textbf{Jetson Nano} in different conditions of the WiFi channel quality. Local processing delay of original models (not displayed): DenseNet-169: $0.27s$, DenseNet-201: $0.42s$, ResNet-152: $0.53s$.}
    \label{fig:delay_barplot}
\end{figure}

First, we analyze the end-to-end delay and power consumption in the configuration with Jetson Nano as mobile device and WiFi as a communication technology. Figure~\ref{fig:delay_barplot} shows the end-to-end average delay with $95\%$ confidence interval using edge offloading, and \FRAMEWORK with $3,6$ and $12$ channels, for each of the four considered models. The results show that in high channel quality regimes, some models work well with edge offloading. However, when the link quality decreases, \FRAMEWORK offers lower average delay and delay variability. We exclude from these graphs MobileNetV2, whose execution time, independent from channel quality, is $0.17s$ (confidence interval is negligible). 

\begin{figure}[t]
    \centering 
    \includegraphics[width=0.95\linewidth]{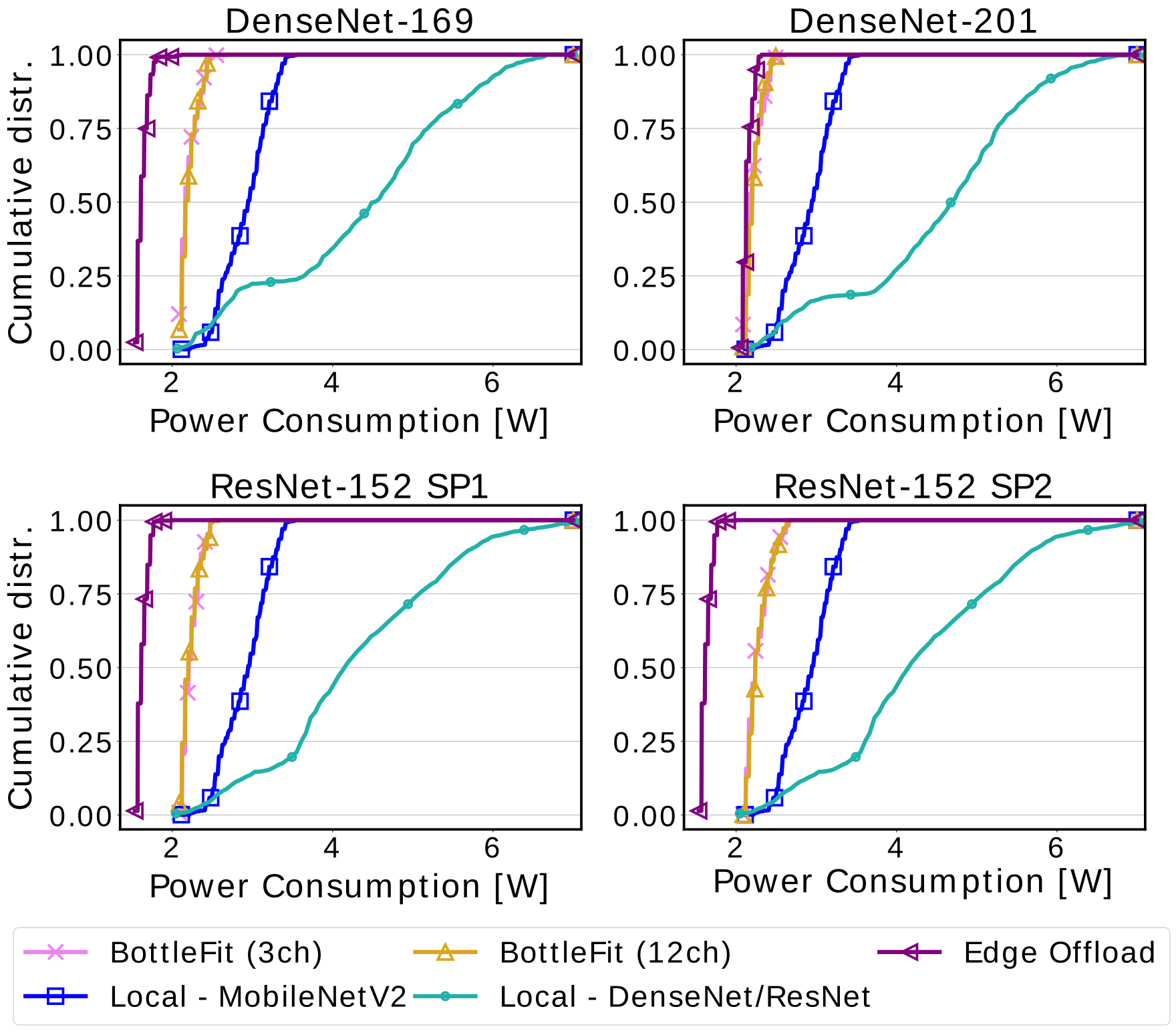}
    \vspace{-0.5em}
    \caption{Cumulative distribution of power consumption for the models considered. We compare edge offloading as well as local processing both using MobileNetV2 and full models (DenseNet/ResNet).\vspace{-0.45cm}}
    \label{fig:energy_plot}
\end{figure}

\begin{figure}[t]
    \centering
    \includegraphics[width=0.93\linewidth]{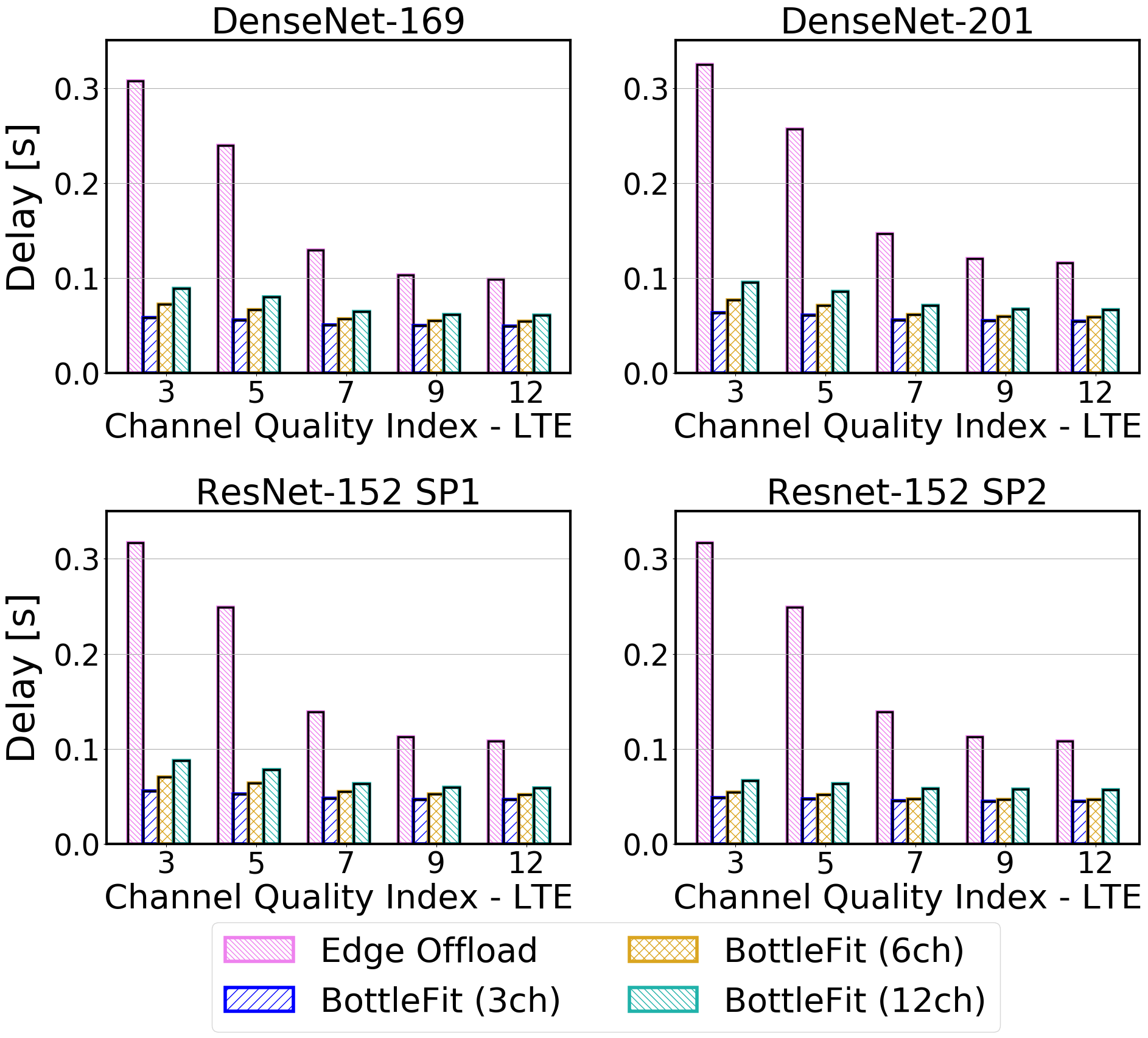}
    \vspace{-0.75em}
    \caption{Delay of the considered models run on \textbf{Jetson Nano} in different LTE Channel Quality Index conditions.}
    \label{fig:delay_LTE_barplot}
\end{figure}

Figure~\ref{fig:energy_plot} depicts the cumulative distribution of the power consumption -- measured at the mobile device -- of the considered models. We first observe that the Local DenseNet/ResNet distributions exhibit peak value well over $6$W, with average value equal to $4.3$W. Being a model designed to be run on mobile devices, MobileNetV2 shows impressive power saving, limiting its range below $4$W (and average $2.9$W). In line with expectations, we see that in all cases except DenseNet-201, offloading to the edge server is the strategy yielding the lowest power used, which never exceeds $2$W (average value $1.6$W).

However, as pointed out in Fig.~\ref{fig:delay_barplot}, the delay offered by edge offloading can exceed the constraint imposed by the application. For this reason, \FRAMEWORK should be preferred, since with power consumption only $37\%$ higher than edge offloading, and $89\%$ lower than local processing, it offers  comparable accuracy when using 12 channels configuration (less than 1\% loss in all cases), while delivering the most stable and low delays. 
Similarly, as shown in  Fig.\ref{fig:delay_LTE_barplot}, when the system employs LTE, we observe a reduction of end-to-end delay in the range $45\%-70\%$ compared to offloading. Conversely to the WiFi configuration, even for high rates, \FRAMEWORK always outperforms offloading.

\subsection{Latency and Power -- Raspberry Pi 4/LoRa}\label{sec:res_raspberry}

\begin{figure}[t]
    \centering
    \includegraphics[width=0.95\linewidth]{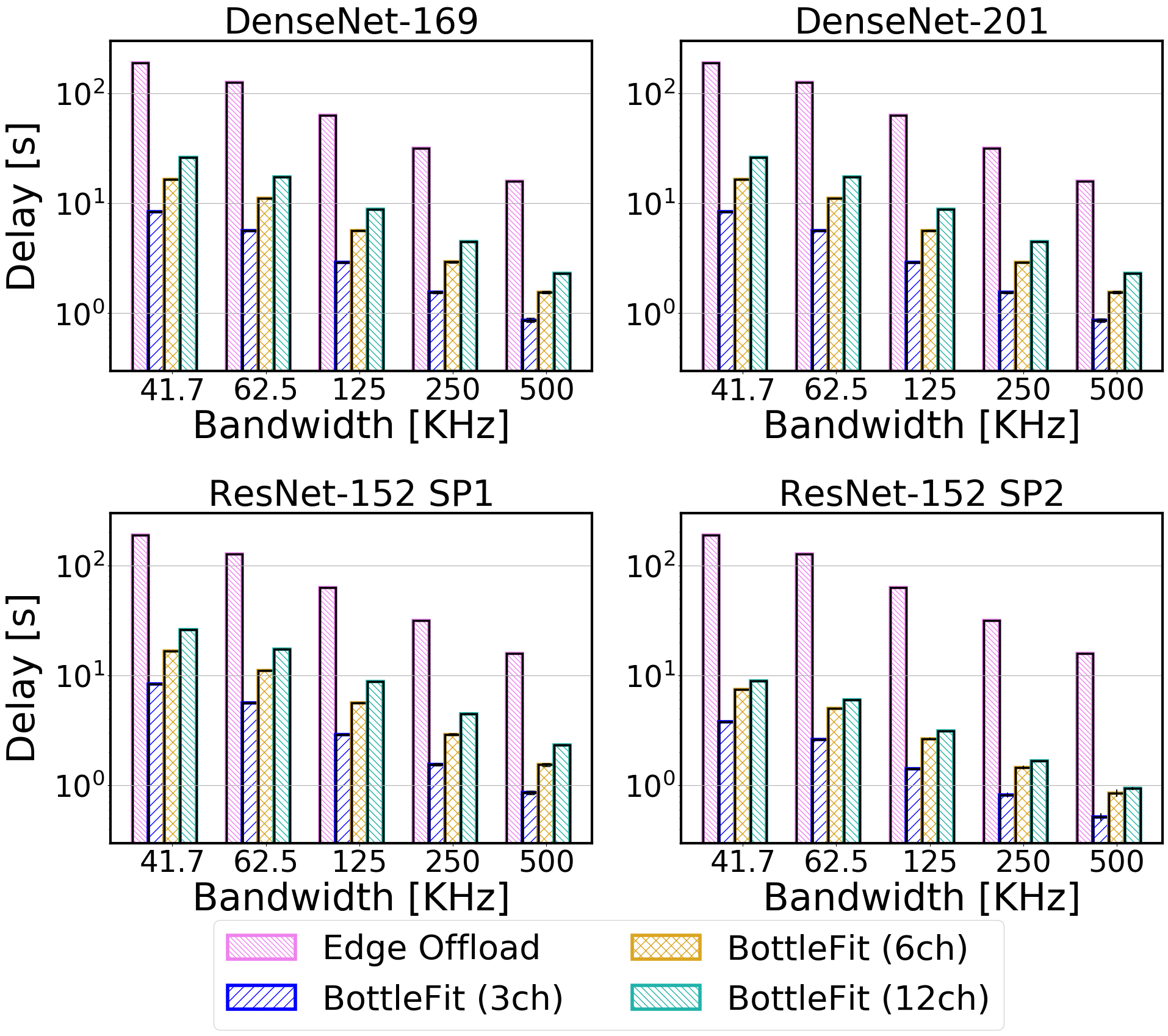}
    \vspace{-0.75em}
    \caption{Delay of the considered models run on \textbf{Raspberry Pi 4} with different LoRa bandwidths. Local processing delay of original models (not displayed): DenseNet-169: $2.34s$, DenseNet-201: $2.55s$, ResNet-152: $4.29s$, MobileNetV2: $1.97s$.\vspace{-1em}}
    \label{fig:delay_raspberry_barplot}
\end{figure}

\begin{figure}[t]
    \centering
    \includegraphics[width=0.9\linewidth]{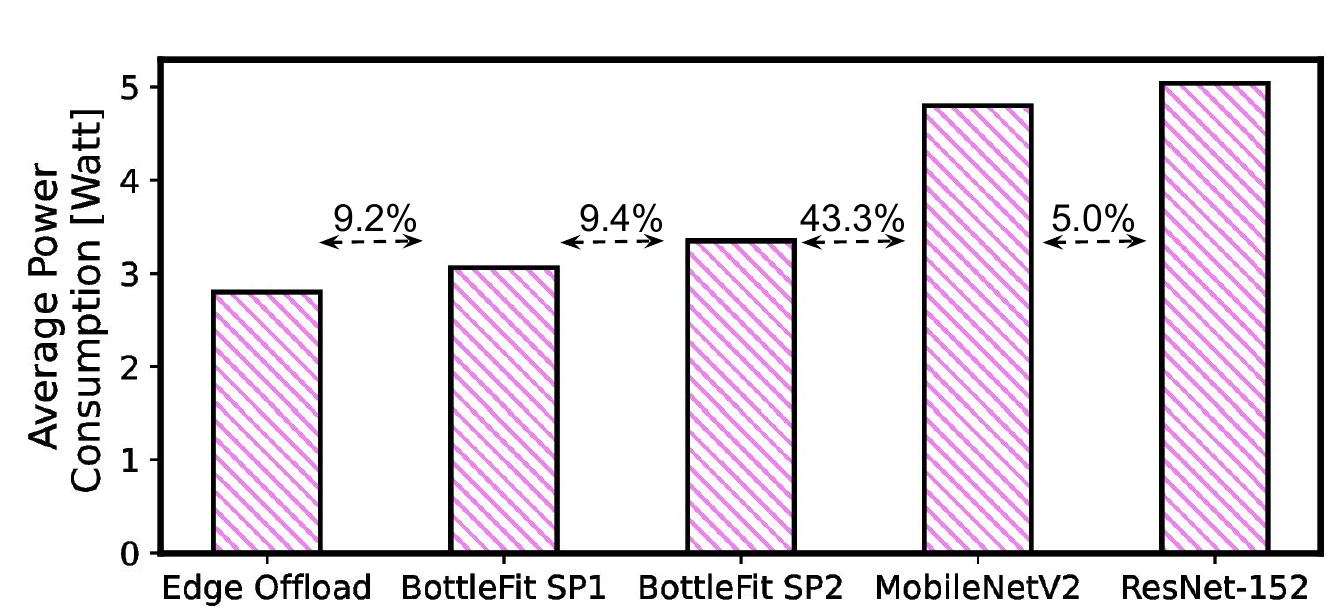}
    \vspace{-0.5em}
    \caption{Average power consumption on Raspberry Pi 4 for the different approaches presented.}
    \label{fig:energy_rpi}
\end{figure}

We present in Figs.~\ref{fig:delay_raspberry_barplot} and \ref{fig:energy_rpi} the delay and power consumption obtained by executing the image classification application on a Raspberry Pi 4 with Long Range (LoRa) connectivity~\cite{adelantado2017understanding}, which is not equipped with a GPU. The purpose of these experiments is to evaluate \FRAMEWORK in the context of a low-power device equipped with low-power, low-rate connectivity. We estimated the network delay by taking the nominal data rates of LoRa with spreading factor 6 and coding rate 4:5 (which yield the highest throughput). As expected, Fig.~\ref{fig:delay_raspberry_barplot} shows that \FRAMEWORK outperforms a full edge offloading approach, given the compression provided by the bottleneck. On the other hand, the networking delay given by LoRa makes local computation more delay-effective.

The advantage of local computing with respect to \FRAMEWORK comes to the detriment of power consumption. Figure~\ref{fig:energy_rpi} shows the average power consumption experienced by the Raspberry Pi 4 as a function of different models. Confidence intervals are not shown as those were below 1\%. \textit{It indicates that \FRAMEWORK presents power consumption comparable to edge offloading, while saving up to 50\% power consumption with respect to local computing approaches.}




\section{Conclusions}
\label{sec:conclusions}

In this paper, we have proposed \FRAMEWORK, a new framework for split computing. We have applied \FRAMEWORK on cutting-edge \gls{dnn} models in image classification, and show that \FRAMEWORK achieves 77.1\% data compression with up to 0.6\% accuracy loss on ImageNet dataset. We experimentally measure the power consumption and latency of an image classification application, and shown that \FRAMEWORK decreases power consumption and latency respectively by up to 49\% and 89\% with respect to local computing and by 37\% and 55\% w.r.t. edge offloading. We also compare \FRAMEWORK with state-of-the-art autoencoders-based approaches, and show that (i) \FRAMEWORK reduces power consumption and execution time respectively by up to 54\% and 62\%; (ii) the size of the head model executed on the mobile device is 83 times smaller.
To achieve more efficient split computing, it would be essential to further improve the tradeoff between transferred data size and model accuracy while keeping encoder lightweight as discussed in~\cite{matsubara2022supervised,matsubara2022sc2}.


\bibliographystyle{IEEEtran}
\bibliography{references}
\end{document}